\documentclass[10pt,twocolumn,letterpaper]{article}

\makeatletter
\renewcommand\paragraph{\@startsection{paragraph}{4}{\z@}{1ex}{-1em}{\normalfont\normalsize\bfseries}}
\makeatother

\usepackage[pagenumbers]{cvpr} 
\makeatletter
\@namedef{ver@everyshi.sty}{}
\makeatother

\usepackage[dvipsnames,svgnames,x11names]{xcolor}
\usepackage{tikz}
\usetikzlibrary{arrows.meta,shapes,calc,matrix,fit,positioning,backgrounds,decorations.markings,fadings}
\usepackage{pgfplots}
\usepackage{pgfplotstable}
\pgfplotsset{compat=1.9}
\usepackage{xstring}

\usepgfplotslibrary{external}

\IfBeginWith*{\jobname}{fig/extern/}{\finalcopy}{}

\tikzstyle{every picture}+=[
	remember picture,
	every text node part/.style={align=center},
	every matrix/.append style={ampersand replacement=\&},
]
\tikzstyle{tight} = [inner sep=0pt,outer sep=0pt]
\tikzstyle{node}  = [draw,circle,tight,minimum size=12pt,anchor=center]
\tikzstyle{op}    = [draw,circle,tight]
\tikzstyle{dot}   = [fill,draw,circle,inner sep=1pt,outer sep=0]
\tikzstyle{pt}    = [fill,draw,circle,inner sep=1.5pt,outer sep=.2pt]
\tikzstyle{box}   = [draw,rectangle,inner sep=3pt]
\tikzstyle{high}  = [black!60]
\tikzstyle{group} = [high,box,opacity=.5]
\tikzstyle{dim1}  = [fill opacity=.3,text opacity=1]
\tikzstyle{dim2}  = [fill opacity=.5,text opacity=1]
\tikzstyle{dim3}  = [fill opacity=.7,text opacity=1]
\tikzstyle{rectc} = [tight,transform shape]
\tikzstyle{rect}  = [rectc,anchor=south west]

\tikzset{every mark/.append style={solid}}
\pgfplotsset{
	grid=both, width=\columnwidth, try min ticks=5,
	every axis/.append style={font=\small},
	every axis plot/.append style={thick,mark=none,mark size=1.8,tension=0.18},
	legend cell align=left, legend style={fill opacity=0.8},
	xticklabel={\pgfmathprintnumber[assume math mode=true]{\tick}},
	yticklabel={\pgfmathprintnumber[assume math mode=true]{\tick}},
	nodes near coords math/.style={
		nodes near coords={\pgfmathprintnumber[assume math mode=true]{\pgfplotspointmeta}},
	},
}

\pgfplotsset{
	dash/.style={mark=o,dashed,opacity=0.6},
	dott/.style={mark=o,dotted,opacity=0.6},
	nolim/.style={enlargelimits=false},
	plain/.style={every axis plot/.append style={},nolim,grid=none},
}

\pgfdeclarelayer{bg4}
\pgfdeclarelayer{bg3}
\pgfdeclarelayer{bg2}
\pgfdeclarelayer{bg1}
\pgfdeclarelayer{fg1}
\pgfdeclarelayer{fg2}
\pgfdeclarelayer{fg3}
\pgfdeclarelayer{fg4}
\pgfsetlayers{bg4,bg3,bg2,bg1,main,fg1,fg2,fg3,fg4}

\pgfkeys{/tikz/.cd, aspect/.store in=\aspect, aspect=1}
\pgfkeys{/tikz/.cd, depth/.store in=\depth, depth=.5}
\pgfkeys{/tikz/.cd, stepx/.store in=\step, stepx=1}

\tikzstyle{geom} = [line join=bevel,aspect=1,depth=.5,z={(\depth*\aspect,\depth)}]
\tikzstyle{wire} = [geom,draw,thick]

\def\cx[#1,#2,#3]{#1}
\def\cy[#1,#2,#3]{#2}
\def\cz[#1,#2,#3]{#3}
\def\ex[#1,#2,#3]{#1,0,0}
\def\ey[#1,#2,#3]{0,#2,0}
\def\ez[#1,#2,#3]{0,0,#3}

\usepackage[dvipsnames]{xcolor}
\usepackage{times}
\usepackage{epsfig}
\usepackage{graphicx}
\usepackage{amsmath}
\usepackage{amssymb}

\usepackage{float}
\usepackage{multirow}
\usepackage[numbers]{natbib}
\usepackage{enumitem}
\usepackage{array,booktabs}
\usepackage{bbm}
\usepackage{colortbl}

\definecolor{cvprblue}{rgb}{0.21,0.49,0.74}
\usepackage[pagebackref,breaklinks,colorlinks,citecolor=cvprblue]{hyperref}

\def\paperID{27} 
\def\confName{CVPR}
\def\confYear{2024}

\title{\Ours: Attention-based pooling for interpretable image recognition}

\newcommand{\eq}[1]{(\ref{eq:#1})}

\newcommand{\Th}[1]{\textsc{#1}}
\newcommand{\mr}[2]{\multirow{#1}{*}{#2}}
\newcommand{\mc}[2]{\multicolumn{#1}{c}{#2}}

\newcommand{\red}[1]{{\textcolor{red}{#1}}}
\newcommand{\blue}[1]{{\textcolor{blue}{#1}}}

\newcommand{\citeme}[1]{\red{[XX]}}
\newcommand{\refme}[1]{\red{(XX)}}

\newcommand{\tran}{^\top}

\newcommand{\real}{\mathbb{R}}

\newcommand{\vect}{\operatorname{vec}}

\newcommand{\softmax}{\operatorname{softmax}}

\newcommand{\defn}{\mathrel{:=}}

\newcommand{\cX}{\mathcal{X}}

\newcommand{\vF}{\mathbf{F}}

\newcommand{\va}{\mathbf{a}}

\newcommand{\vf}{\mathbf{f}}

\newcommand{\vp}{\mathbf{p}}
\newcommand{\vq}{\mathbf{q}}

\newcommand{\vx}{\mathbf{x}}
\newcommand{\vy}{\mathbf{y}}

\newcommand{\valpha}{{\boldsymbol{\alpha}}}

\newcommand{\vphi}{{\boldsymbol{\phi}}}

\makeatletter
\newcommand*\bdot{\mathpalette\bdot@{.7}}
\newcommand*\bdot@[2]{\mathbin{\vcenter{\hbox{\scalebox{#2}{$\m@th#1\bullet$}}}}}
\makeatother

\makeatletter
\DeclareRobustCommand\onedot{\futurelet\@let@token\@onedot}
\def\@onedot{\ifx\@let@token.\else.\null\fi\xspace}

\def\eg{\emph{e.g}\onedot}

 \def\vs{\emph{vs}\onedot}

\makeatother

\newcommand{\gap}{{\textsc{gap}}\xspace}

\newcommand{\ca}{{\textsc{ca}}\xspace}
\newcommand{\sa}{{\textsc{sa}}\xspace}

\newcommand{\cls}{{\textsc{cls}}\xspace}
\newcommand{\our}{{\textsc{ca}}}
\newcommand{\ours}{{\textsc{ca}}\xspace}

\newcommand{\Ours}{{CA-Stream}\xspace}
\newcommand{\OURS}{{Cross-Attention Stream}\xspace}
\newcommand{\PO}{{\textsc{proj}$\to$\our}\xspace}

\newcommand{\gain}[1]{}

\definecolor{TableColor}{rgb}{0.835, 0.894, 0.968}

 \author{Felipe Torres$^1$, Hanwei Zhang$^2$, Ronan Sicre$^1$, St\'ephane Ayache$^1$, Yannis Avrithis$^3$\\
$^1$Centrale Marseille, Aix Marseille Univ, CNRS, LIS, France\\$^2$Institute of  Intelligent Software, China\\
$^3$Institute of Advanced Research on Artificial Intelligence (IARAI), Austria\\
{\tt\small \{felipe.torres,ronan.sicre,stephane.ayache\}@lis-lab.fr},\\ {\tt\small zhanghanwei0912@gmail.com}, {\tt\small yannis@avrithis.net}
 }

\begin{document}
\maketitle
\begin{abstract}
    Explanations obtained from transformer-based architectures in the form of raw attention, can be seen as a class-agnostic saliency map. Additionally, attention-based pooling serves as a form of masking the in feature space. Motivated by this observation, we design an attention-based pooling mechanism intended to replace Global Average Pooling (\gap) at inference. This mechanism, called \emph{\OURS (\Ours)}, comprises a stream of cross attention blocks interacting with features at different network depths. \Ours enhances interpretability in models, while preserving recognition performance.
 \end{abstract}
\section{Introduction}
\label{sec:intro}

\emph{Convolutional neural networks} (CNN) have attained tremendous success in computer vision~\cite{he2016deep, liu2022convnet}, but interpreting their predictions remains challenging. Most explanations are based on saliency maps, using methods derived from \emph{class activation mapping} (CAM). \emph{Vision transformers}~\cite{dosovitskiy2020image} are now strong competitors of convolutional networks, characterized by global interactions between patch embeddings in the form of \emph{self attention}. Based on the classification (\cls) token, an explanation map in the form of \emph{raw attention} can be constructed. However, these maps are class-agnostic, often of low quality~\cite{dino}, and dedicated interpretability methods are required to explain models~\cite{chefer2021transformer}.

In CNNs, features are pooled into a global representation by \emph{global average pooling} (GAP). In transformers, a global representation is obtained by cross-attention between patch embeddings and the \cls token. In this work, we make a connection between CAM-based saliency maps and raw attention from the \cls token, observing that attention-based pooling is a form of \emph{masking in the feature space}.
Motivated by this observation, we design a pooling mechanism that generates a global representation to be used at inference, replacing GAP and improving interpretability.

Our approach, called \emph{\OURS} (\emph{\Ours}), consists of a branch in parallel with the backbone network, allowing interactions between feature maps and the \cls token through cross-attention at different stages of the network.
The \cls token embedding is a learnable parameter and, at the output of the stream, provides a global image representation for classification.


More specifically, we make the following contributions:
\begin{enumerate}[itemsep=0pt, parsep=0pt, topsep=1pt]
    \item We demonstrate that attention-based pooling in vision transformers is the same as soft masking by a class-agnostic CAM-based saliency map (\autoref{subsec:motiv}).
    \item We design an attention-based pooling mechanism, inject it in convolutional networks to replace GAP and study its effect on post-hoc interpretability (\autoref{subsec:CA-base}). 
    \item We show improved explanations for a trained model and provides a class-agnostic raw attention map (\autoref{sec:exp}).
\end{enumerate}

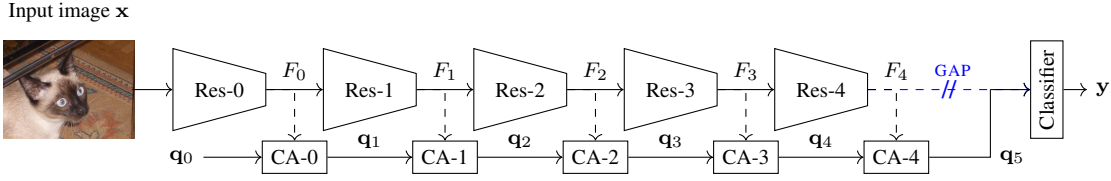
\begin{figure*}
\centering
\begin{tikzpicture}[
	font={\footnotesize},
	trap/.style={trapezium, rotate=-90,trapezium angle=75},
]
	\node(input) at (-5.5, 0) {\includegraphics[width=.1\textwidth]{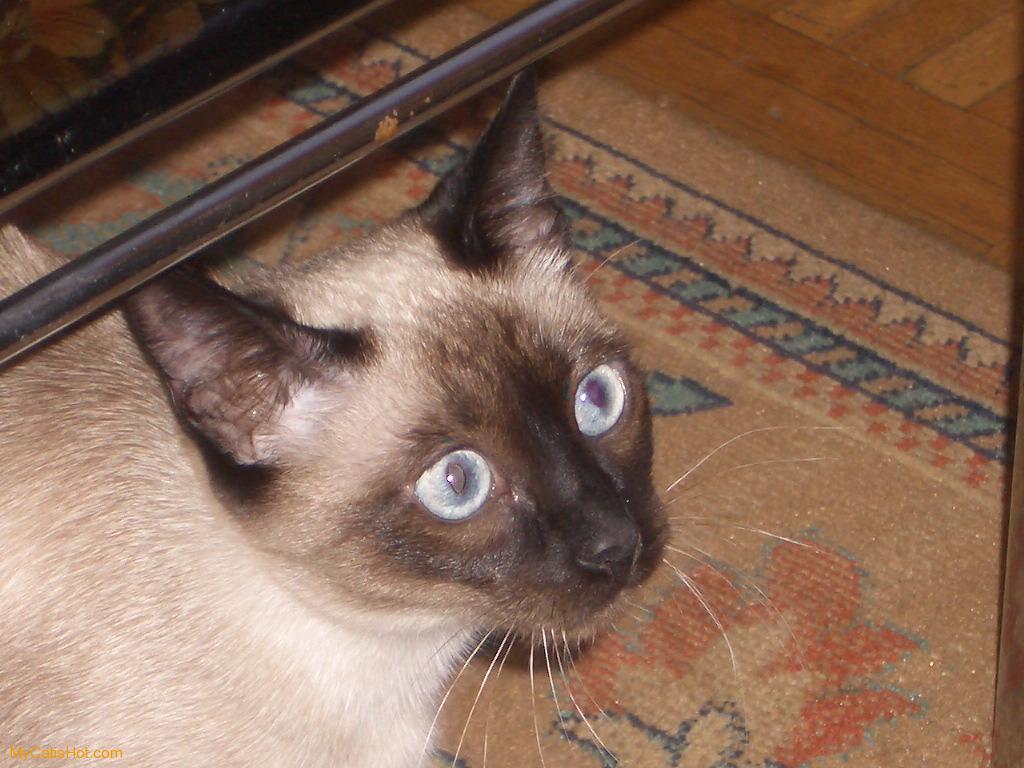}};
	\node[above] at (input.north) {Input image $\vx$};
	\node[draw, trap] (res0) at (-3.5,0) {\rotatebox{90}{\parbox{1.0cm}{\centering{Res-0}}}};
	\node[draw, trap] (res1) at (-1.5,0) {\rotatebox{90}{\parbox{1.0cm}{\centering{Res-1}}}};
	\node[draw, trap] (res2) at (0.5,0) {\rotatebox{90}{\parbox{1.0cm}{\centering{Res-2}}}};
	\node[draw, trap] (res3) at (2.5,0) {\rotatebox{90}{\parbox{1.0cm}{\centering{Res-3}}}};
	\node[draw, trap] (res4) at (4.5,0) {\rotatebox{90}{\parbox{1.0cm}{\centering{Res-4}}}};
	\node[](empt1) at (6.75, 0){};
	\node[draw, rotate=90, align=center] (class) at (7.5,0) {Classifier};
	\node(logit) at (8.25, 0) {$\vy$};
	\node[](clsin) at (-4, -0.9) {{$\vq_0$}};
	\node[draw](CA0) at (-2.5, -0.9) {{CA-0}};
	\node[draw](CA1) at (-0.5, -0.9) {{CA-1}};
	\node[draw](CA2) at (1.5, -0.9)  {{CA-2}};
	\node[draw](CA3) at (3.5, -0.9)  {{CA-3}};
	\node[draw](CA4) at (5.5, -0.9)  {{CA-4}};

	\node(empt0) at (-4.65, 0) {};
	\draw[->] (empt0.center) -- node {} (res0);
	\draw[->] (res0) -- node[above] {$F_0$} (res1);
	\draw[->] (res1) -- node[above] {$F_1$} (res2);
	\draw[->] (res2) -- node[above] {$F_2$} (res3);
	\draw[->] (res3) -- node[above] {$F_3$} (res4);
	\draw[->, blue, dashed] (res4) -- node {\blue{\normalsize//}} (class);
	\node[](GAP) at (6.25,0.25) {\blue{$\gap$}};
	\draw[->] (class) -- node {} (logit);
	\draw[->] (clsin) -- node {} (CA0);
	\draw[dashed, ->] (res0.north) -|node {} (CA0);
	\draw[->] (CA0) -- node[above] {$\vq_1$} (CA1);
	\draw[dashed, ->] (res1.north) -|node {} (CA1);
	\draw[->] (CA1) -- node[above] {$\vq_2$} (CA2);
	\draw[dashed, ->] (res2.north) -|node {} (CA2);
	\draw[->] (CA2) -- node[above] {$\vq_3$} (CA3);
	\draw[dashed, ->] (res3.north) -|node {} (CA3);
	\draw[->] (CA3) -- node[above] {$\vq_4$} (CA4);
	\draw[dashed, ->] (res4.north) -|node[above] {$F_4$} (CA4);
	\draw[-] (CA4.east) -| node[right] {$\vq_5$} (empt1.center);
	\draw[->] (empt1.center) -- node {} (class);
\end{tikzpicture}
\caption{\emph{\OURS (\Ours\OURS (\Ours) applied to ResNet-based architectures.} Given a network $f$, we replace global average pooling (\gap) by a learned, attention-based pooling mechanism implemented as a stream in parallel to $f$. The feature tensor $F_\ell \in \real^{p_\ell \times d_\ell}$ (\emph{key}) obtained by stage Res-$\ell$ of $f$  interacts with a \cls token (\emph{query}) embedding $\vq_\ell \in \real^{d_\ell}$ in block CA-$\ell$, which contains cross attention~\eq{CA} followed by a linear projection~\eq{qk-layer} to adapt to the dimension of $F_{\ell+1}$. Here, $p_\ell$ is the number of patches (spatial resolution) and $d_\ell$ the embedding dimension. The query is initialized by a learnable parameter $\vq_0 \in \real^{d_0}$, while the output $\vq_5$ of the last cross attention block is used as a global image representation into the classifier.}
\label{fig:fig_method}
\end{figure*}

\section{Related work}

Deep neural networks interpretability is investigated though \emph{Post-hoc interpretability} or \emph{Transparency} ~\citep{lipton18, guidotti2018survey, zhang2021survey}.

\paragraph{Post-hoc interpretability} considers the model as a black-box and provides explanations based on input and output observations. 
These methods can be grouped into sets of possibly overlapping categories. \emph{Gradient-based methods}~\citep{adebayo2018local, springenberg2014striving, baehrens2010explain, simonyan2013deep, smilkov2017smoothgrad, bach2015pixel, sundararajan2017axiomatic} use gradient information to visualize the contribution of different input regions in an image. \emph{CAM-based methods}~\citep{DBLP:journals/corr/abs-1910-01279, DBLP:journals/corr/abs-1710-11063, DBLP:journals/corr/SelvarajuDVCPB16, fu2020axiom, jiang2021layercam, ramaswamy2020ablation} compute saliency maps as a linear combination of feature maps to highlight salient regions in the input image. \emph{Occlusion or masking-based methods}~\citep{petsiuk2018rise, fong2017interpretable, fong2019understanding, schulz2020restricting, ribeiro2016should} instead compute saliency maps based on the prediction changes induced by masking the input image. Finally, \emph{learning-based methods}~\citep{chang2018explaining, dabkowski2017real, phang2020investigating, zolna2020classifier, schulz2020restricting} learn additional models or branches to produce explanations for a given input.

\paragraph{Transparency} modifies the model or its training process to explain it. These approaches are grouped according to the nature of the explanation they provide. \emph{Rule-based methods} ~\citep{wu2018beyond, wu2020regional} approximate the model using a decision tree as a proxy. \emph{Hidden semantic-methods} ~\citep{bau2017network, zhou2018interpreting, zhang2018interpretable, zhou2014object} learn disentangled semantics following a hierarchical structure or object-level concepts. \emph{Prototype-based methods} learn prototypes seen in training images to explain models from intermediate representations. \emph{Attribution-based methods}~\citep{ismail2021improving, Zhou_2022_BMVC, ross2017right, ghaeini2019saliency} propose modifications to the network or its training process, improving interpretable properties of post-hoc attribution methods. Finally, saliency-guided training ~\cite{ismail2021improving, lee2021lfi} design and train a model that aligns images with their saliency based masks during training enhancing recognition and interpretability properties.

Our approach aligns with attribution-based methods. Specifically, we introduce a learnable cross-attention stream, producing a representation that replaces \gap.

\paragraph{Attention-based architectures}
Attention is a mechanism introduced into convolutional neural networks to enhance their recognition capabilities ~\citep{bello2019attention, ramachandran2019stand, shen2020global}. Following the success of vision transformers (ViT) ~\citep{dosovitskiy2020image}, fully attention-based architectures are now competitive with  convolutional neural networks,  while drawing inspiration from them to enhance their recognition capabilities ~\cite{graham2021levit,xiao2021early, liu2021swin, heo2021rethinking}.

Unlike similar approaches combining ideas from convolutions in transformers~\citep{peng2021conformer, li2021scouter, touvron2021augmenting}, we propose to add an attention-based pooling mechanism  in convolutional models, enhancing post-hoc interpretability properties without degrading classification accuracy.

\section{Method}
\begin{figure*}[t]
\scriptsize
\centering
\setlength{\tabcolsep}{1.5pt}
\begin{tabular}{ccccccccc}
	{}&\multirow{2}{*}{Input image}&\multirow{2}{*}{Raw Attention}&\multicolumn{2}{c}{Grad-CAM}&\multicolumn{2}{c}{Grad-CAM++}&\multicolumn{2}{c}{Score-CAM}\\
	{}&{}&{}&GAP&\Ours&GAP&\Ours&GAP&\Ours\\

	{\rotatebox{90}{\tiny CRT screen}}&\multicolumn{1}{c}{\includegraphics[width=0.115\textwidth]{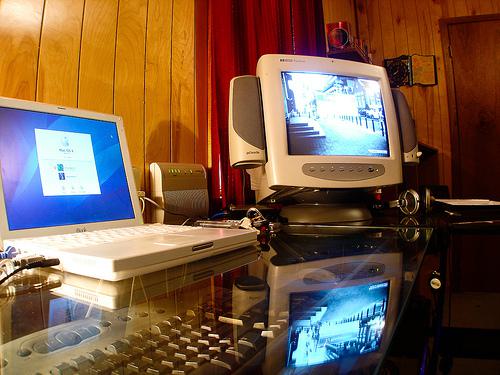}}&\multicolumn{1}{c}{\includegraphics[width=0.115\textwidth]{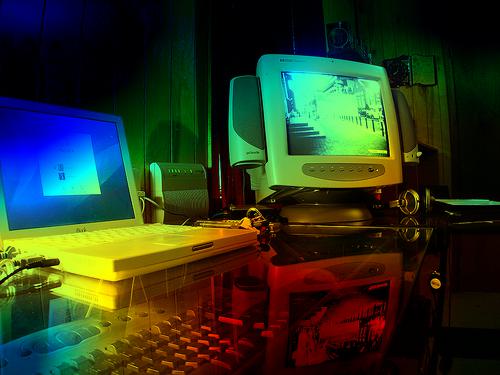}}&\multicolumn{1}{c}{\includegraphics[width=0.115\textwidth]{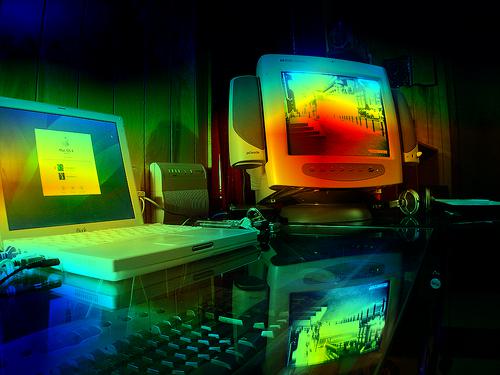}}&\multicolumn{1}{c}{\includegraphics[width=0.115\textwidth]{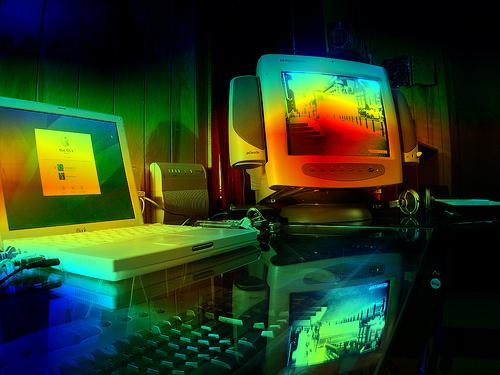}}&\multicolumn{1}{c}{\includegraphics[width=0.115\textwidth]{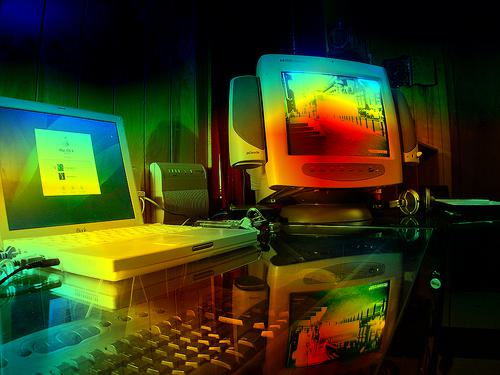}}&\multicolumn{1}{c}{\includegraphics[width=0.115\textwidth]{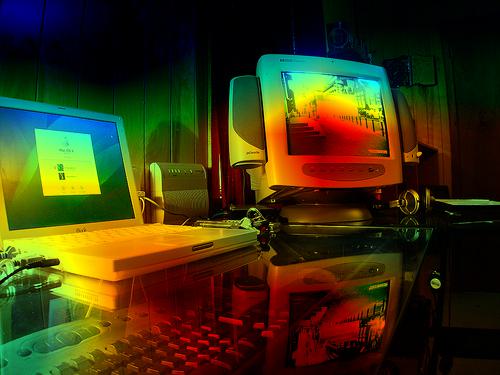}}&\multicolumn{1}{c}{\includegraphics[width=0.115\textwidth]{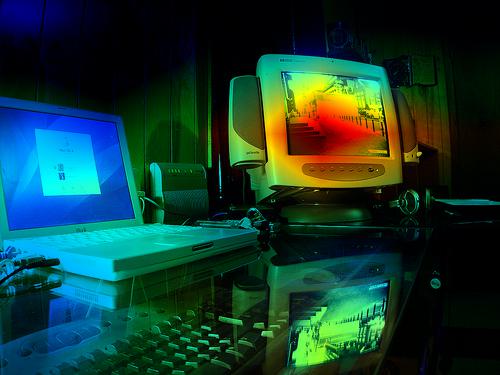}}&\multicolumn{1}{c}{\includegraphics[width=0.115\textwidth]{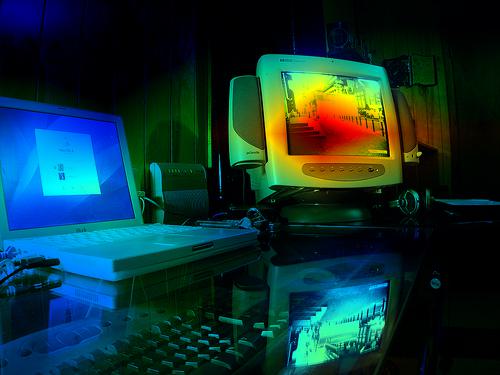}}\\ 

    \textbf   {\rotatebox{90}{\tiny Snowboard}}&\includegraphics[width=0.115\textwidth]{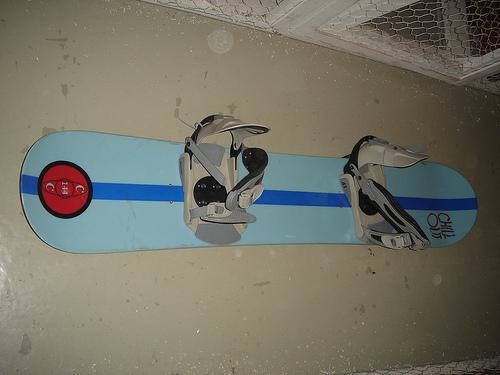}&\includegraphics[width=0.115\textwidth]{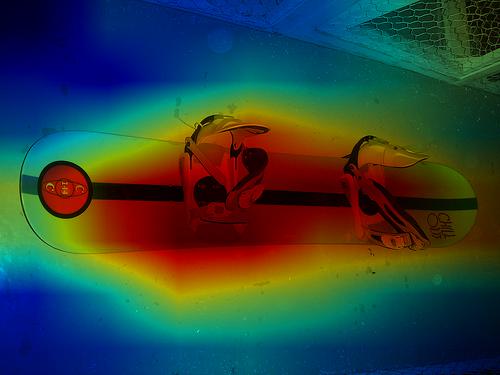}&\includegraphics[width=0.115\textwidth]{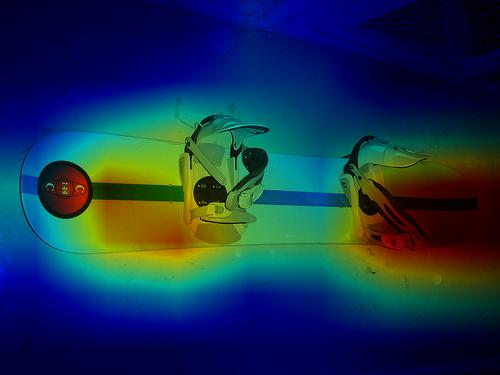}&\includegraphics[width=0.115\textwidth]{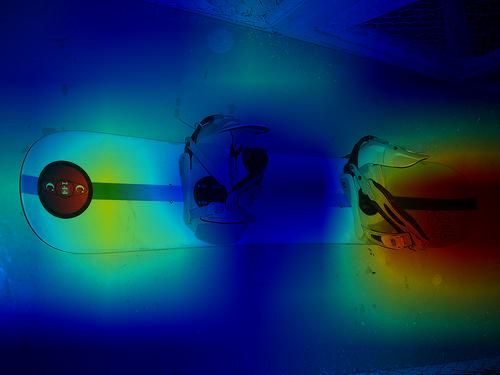}&\includegraphics[width=0.115\textwidth]{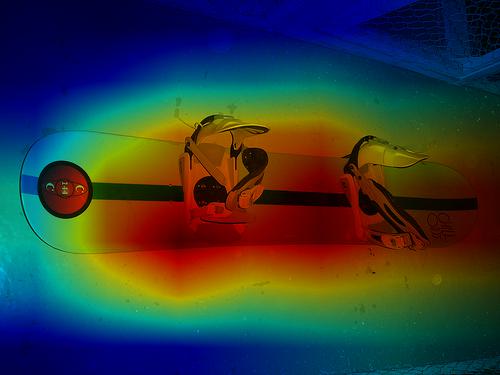}&\includegraphics[width=0.115\textwidth]{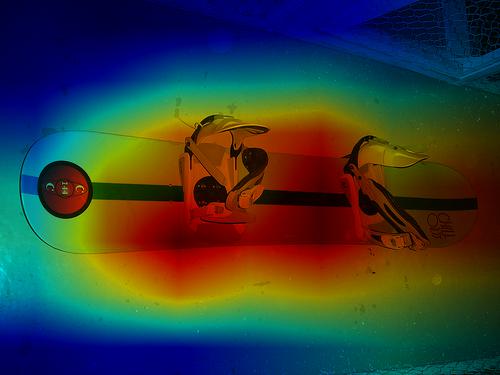}&\includegraphics[width=0.115\textwidth]{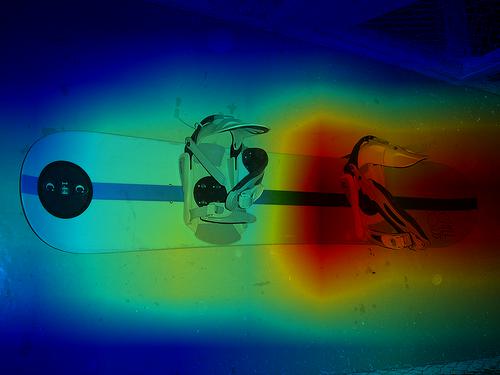}&\includegraphics[width=0.115\textwidth]{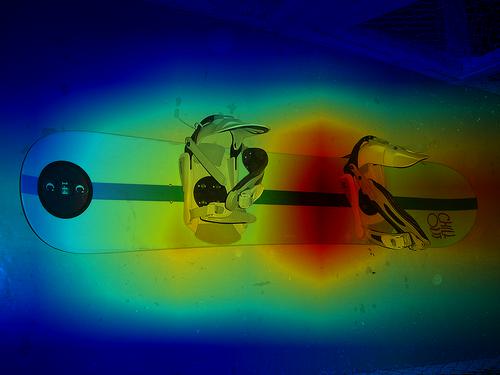}\\
 
\end{tabular}
\caption{Comparison of saliency maps generated by different CAM-based methods, using GAP and our \Ours, on ImageNet images. The raw attention is the one used for pooling by \Ours.}
\label{fig:compmethods}
\end{figure*}
\subsection{Preliminaries and background}
\paragraph{Notation}
\label{subsec:prelim}
Let $f: \cX \rightarrow \real^C$ be a classifier network that maps an input image $\vx \in \cX$ to a logit vector $\vy= f(\vx) \in \real^C$, where $\cX$ is the image space and $C$ is the number of classes. A class probability vector is obtained by $\vp = \softmax(\vy)$. The logit and probability of class $c$ are respectively denoted by $y^c$ and $p^c = \softmax(\vy)^c$. Let $\vF_\ell \in \real^{w_\ell \times h_\ell \times d_\ell}$ be the feature tensor at layer $\ell$ of the network, where $w_\ell \times h_\ell$ is the spatial resolution and $d_\ell$ the embedding dimension, or number of channels. The feature map of channel $k$ is denoted by $F^k_\ell \in \real^{w_\ell \times h_\ell}$.
\paragraph{CAM-based saliency maps}
Given a class of interest $c$ and a layer $\ell$, we consider the saliency maps $S^c_\ell \in \real^{w_\ell \times h_\ell}$ given by the general formula
\begin{equation}
	S^c_\ell \defn h \left( \sum_k \alpha^c_k F^k_\ell \right),
    \label{eq:sal}
\end{equation}
where $\alpha^c_k$ are weights defining a linear combination over channels and $h$ is an activation function. Assuming \emph{global average pooling} (GAP) of the last feature tensor $\vF_L$ followed by a linear classifier, CAM~\citep{zhou2016learning} is defined for the last layer $L$ only, with $h$ being the identity mapping and $\alpha^c_k$ the classifier weight connecting channel $k$ with class $c$.
\paragraph{Self-attention}
Let $X_\ell \in \real^{t_\ell \times d_\ell}$ denote the sequence of token embeddings of a vision transformer~\cite{dosovitskiy2020image} at layer $\ell$, where $t_\ell \defn w_\ell h_\ell + 1$ is the number of tokens, including patch tokens and the \cls token, and $d_\ell$ is the embedding dimension. The \emph{attention matrix} $A \in \real^{t_\ell \times t_\ell}$ expresses pairwise dot-product similarities between queries ($Q$) and keys ($K$), normalized by softmax over rows:
\begin{equation}
	A = \softmax \left( \frac{Q K\tran}{\sqrt{d_\ell}} \right).
    \label{eq:attention}
\end{equation}
For each token, the \emph{self-attention} operation is then defined as an average of all values ($V$) weighted by attention $A$:
\begin{equation}
	\sa(X_\ell) \defn A V \in \real^{t_\ell \times d_\ell}.
    \label{eq:SA}
\end{equation}
At the last layer $L$, the \cls token embedding is used as a global representation for classification as it gathers information from all patches by weighted averaging, replacing \gap. Thus, at the last layer, it is only cross-attention between \cls and the patch tokens that matters.
\subsection{Motivation}
\paragraph{Cross-attention}
\label{subsec:motiv}
Let matrix $F_\ell \in \real^{p_\ell \times d_\ell}$ be a reshaping of feature tensor $\vF_\ell$ at layer $\ell$, where $p_\ell \defn w_\ell h_\ell$ is the number of patch tokens without \cls, and let $\vq_\ell \in \real^{d_\ell}$ be the \cls token embedding at layer $\ell$. By focusing on the \emph{cross-attention} only between the \cls (query) token $\vq_\ell$ and the patch (key) tokens $F_\ell$, attention $A$~\eq{attention} is now a $1 \times p_\ell$ matrix that can be written as a vector $\va \in \real^{p_\ell}$
\begin{equation}
	\va = A\tran = \softmax \left( \frac{F_\ell \vq_\ell}{\sqrt{d_\ell}} \right).
    \label{eq:cross-attention}
\end{equation}
Here, $F_\ell \vq_\ell$ expresses the pairwise similarities between the global \cls feature $\vq_\ell$ and the local patch features $F_\ell$. Now, by replacing $\vq_\ell$ by an arbitrary vector $\valpha \in \real^{d_\ell}$ and writing the feature matrix as $F_\ell = (\vf_\ell^1 \dots \vf_\ell^{d_\ell})$, attention \eq{cross-attention} becomes
\begin{equation}
	\va = h_\ell (F_\ell \valpha) =
		h_\ell \left( \sum_k \alpha_k \vf_\ell^k \right).
    \label{eq:connection}
\end{equation}
This takes the same form as~\eq{sal}, with feature maps $F_\ell^k$ vectorized as $\vf_\ell^k$ and the activation function defined as $h_\ell(\vx) = \softmax(\vx / \sqrt{d_\ell})$.  We thus observe the following.

\emph{Pairwise similarities between one query and all patch token embeddings in cross-attention are the same as a linear combination of feature maps in CAM-based saliency maps.}

One difference between~\eq{sal} and~\eq{connection} is that~\eq{connection} is class-agnostic, although it could be extended by using one query vector per class. For simplicity, we choose the class-agnostic form. We also choose to have no query/key projections. However, we do provide additional experiments
in the appendix.

\paragraph{Pooling or masking}
We integrate an attention mechanism into a network such that making a prediction and explaining it are inherently connected. In particular, considering cross-attention only between \cls and patch tokens~\eq{cross-attention}, equation~\eq{SA} becomes
\begin{align}
	\ca_\ell(\vq_\ell, F_\ell) \defn F_\ell\tran \va = F_\ell\tran h_\ell(F_\ell \vq_\ell) \in \real^{d_\ell}.
    \label{eq:CA}
\end{align}
By writing the transpose of feature matrix as $F_\ell\tran = (\vphi_\ell^1 \dots \vphi_\ell^{p_\ell})$ where $\vphi_\ell^i \in \real^{d_\ell}$ is the feature of patch $i$, this is a weighted average of the local patch features $F_\ell\tran$ with attention vector $\va = (a_1, \dots, a_{p_\ell})$ expressing the weights:
\begin{align}
	\ca_\ell(\vq_\ell, F_\ell) \defn F_\ell\tran \va = \sum_i a_i \vphi_\ell^i.
    \label{eq:CA-gap}
\end{align}
We can think of it as as feature \emph{reweighting} or \emph{soft masking} in the feature space, followed by \gap.

Now, considering that $\va$ is obtained exactly as CAM-based saliency maps~\eq{connection}, this operation is similar to occlusion (masking)-based methods~\citep{petsiuk2018rise, fong2017interpretable, fong2019understanding, schulz2020restricting, ribeiro2016should,DBLP:journals/corr/abs-1910-01279, zhang2023opti} and evaluation metrics~\cite{DBLP:journals/corr/abs-1710-11063, petsiuk2018rise}, where a CAM-based saliency map is commonly used to mask the input image. We thus observe the following.

	\emph{Attention-based pooling is a form of feature reweighting or soft masking in the feature space followed by \gap, where the weights are given by a class-agnostic CAM-based saliency map.}

\subsection{Cross-attention stream}
\label{subsec:CA-base}
Motivated by these observations, we design a \emph{\OURS} (\emph{\Ours}) in parallel to any network. It takes input features at key locations of the network and uses cross-attention to build a global image representation and replace $\gap$ before the classifier. An example is shown in \autoref{fig:fig_method}, applied to a ResNet-based architecture.

\paragraph{Architecture}

More formally, given a network $f$, we consider points between blocks of $f$ where critical operations take place, such as change of spatial resolution or embedding dimension, \eg between residual blocks on ResNet. We decompose $f$ at these points as
\begin{equation}
	f = g \circ \gap \circ f_L \circ \dots \circ f_0
    \label{eq:f-decomp}
\end{equation}
such that features $F_\ell \in \real^{p_\ell \times d_\ell}$ of layer $\ell$ are initialized as $F_{-1} = \vx$ and updated according to
\begin{equation}
	F_\ell = f_\ell(F_{\ell-1})
    \label{eq:f-layer}
\end{equation}
for $0 \le \ell \le L$, where $p_\ell$ is the number of patch tokens and $d_\ell$ the embedding dimension of stage $\ell$. The last layer features $F_L$ are followed by \gap and $g: \real^{d_L} \to \real^C$ is the classifier, mapping to the logit vector $\vy$.

In parallel, we initialize a classification token embedding as a learnable parameter $\vq_0 \in \real^{d_0}$ and we build a sequence of updated embeddings $\vq_\ell \in \real^{d_\ell}$ along a stream that interacts with $F_\ell$ at each stage $\ell$. Referring to the global representation $\vq_\ell$ as \emph{query} or \cls and to the local image features $F_\ell$ as \emph{key} or patch embeddings, the interaction consists of cross-attention followed by a linear projection $W_\ell \in \real^{d_{\ell+1} \times d_\ell}$ to account for changes of embedding dimension between the corresponding stages of $f$:
\begin{equation}
	\vq_{\ell+1} = W_\ell \cdot \ca_\ell(\vq_\ell, F_\ell),
    \label{eq:qk-layer}
\end{equation}
for $0 \le \ell \le L$, where $\ca_\ell$ is defined as in~\eq{CA}.

Image features do not change by injecting our \Ours into network $f$. However, the final global image representation does, hence the prediction does too. In particular, at the last stage $L$, $\vq_{L+1}$ is used as a global image representation for classification, replacing \gap over $F_L$. Therefore,  final prediction is $g(\vq_{L+1}) \in \real^C$. Unlike \gap, the weights of different image patches in the linear combination are non-uniform, enhancing the contribution of relevant patches in the prediction.


\paragraph{Training}

The network $f$ is pretrained and remains frozen while we learn the parameters of our \Ours on the same training set as $f$. The classifier is kept frozen too. Referring to~\eq{f-decomp}, $f_0, \dots, f_L$ and $g$ are fixed, while \gap is replaced by learned weighted averaging, with the weights obtained by the \Ours.

\paragraph{Inference}

As it stands, the \Ours is an addition to the baseline architecture, which enhances the interpretability properties of a model. We thus investigate interpretability using CAM-based methods on both baseline \gap and \Ours in the following section.

\section{Experiments}
\label{sec:exp}
\paragraph{Experimental setup}
We train and evaluate our models on the ImageNet ILSVRC-2012 dataset~\cite{deng2009imagenet}, using the training and validation sets respectively. We experiment on pretrained and frozen ResNets~\cite{he2016deep} and ConvNeXt~\cite{liu2022convnet} models and provide more details in the appendix. We measure the interpretability properties of our approach by first generating saliency maps employing existing methods based on CAM (Grad-CAM~\cite{DBLP:journals/corr/SelvarajuDVCPB16}, Grad-CAM++~\cite{DBLP:journals/corr/abs-1710-11063}, ScoreCAM~\cite{DBLP:journals/corr/abs-1910-01279}) with and without \Ours. Then, following~\cite{zhang2023opti}, we compute changes in the predictive power of  a masked image measured by \emph{average drop} (AD)~\cite{DBLP:journals/corr/abs-1710-11063} and \emph{average gain} (AG)~\cite{zhang2023opti}, the proportion of better explanations measured by \emph{average increase} (AI)~\cite{DBLP:journals/corr/abs-1710-11063} and finally the impact of different extents of masking via \emph{insertion} (I) and \emph{deletion} (D)~\citep{petsiuk2018rise}.

\paragraph{Qualitative results}

In \autoref{fig:compmethods}, we show saliency maps obtained using either \gap and \ours, as well as the raw attention representation from \Ours. We observe that CAM-based attributions obtained using our \ours are similar to those generated with \gap. We expect this behaviour as we do not modify the model or the weighting coefficients. Since raw attention is class-agnostic, it can be used to gain insight on what the model attends to in unseen data. We iterate upon this in the appendix.

\paragraph{Quantitative evaluation}
\begin{table}[t]
    \centering
    \scriptsize
    \setlength{\tabcolsep}{2.5pt}
    \renewcommand{\arraystretch}{1.025}
    \begin{tabular}{llcccccc}
        \toprule
        \Th{Network}&\Th{Pooling}&\mc{2}{}&\mc{2}{}&\Th{Acc$\uparrow$}\\\midrule
        \mr{2}{ResNet-50}&\gap&\mc{2}{}&\mc{2}{}&74.55\\
        &\ours&\mc{2}{}&\mc{2}{}&74.70\\\midrule
        \mr{2}{ConvNeXt-B}&\gap&\mc{2}{}&\mc{2}{}&83.72\\
        &\ours&\mc{2}{}&\mc{2}{}&83.51\\\midrule
        \Th{Network}&\Th{Attribution}&\Th{Pooling}&\Th{AD$\downarrow$}&\Th{AG$\uparrow$}&\Th{AI$\uparrow$}&\Th{I$\uparrow$}&\Th{D$\downarrow$}\\\midrule
	    \mr{7}{\Th{ResNet-50}}&\mr{2}{Grad-CAM}&\gap&13.04&17.56&44.47&72.57&\textbf{13.24}\\ %
		& &\ours&\textbf{12.54}&\textbf{22.67}&\textbf{48.56}&\textbf{75.53}&13.50\\\cmidrule{2-8} %
		& \mr{2}{Grad-CAM++}&\gap&\textbf{13.79}&15.87&42.08&72.32&\textbf{13.33}\\ %
		& &\ours&13.99&\textbf{19.29}&\textbf{44.60}&\textbf{75.21}&13.78\\\cmidrule{2-8} %
		& \mr{2}{Score-CAM}&\gap&8.83&17.97&48.46&71.99&\textbf{14.31}\\ %
		& &\ours&\textbf{7.09}&\textbf{23.65}&\textbf{54.20}&\textbf{74.91}&14.68\\\midrule
	    \mr{7}{\Th{ConvNeXt-B}}&\mr{2}{Grad-CAM}&\gap&33.72&2.43&15.25&52.85&\textbf{29.57}\\ 
		& &\ours&\textbf{19.45}&\textbf{13.96}&\textbf{32.89}&\textbf{86.38}&45.29\\\cmidrule{2-8} 
		& \mr{2}{Grad-CAM++}&\gap&\textbf{34.01}&2.37&15.60&52.83&\textbf{29.17}\\ 
		& &\ours&36.69&\textbf{8.00}&\textbf{21.95}&\textbf{85.39}&53.42\\\cmidrule{2-8} 
		& \mr{2}{Score-CAM}&\gap&43.55&2.23&15.67&50.96&\textbf{39.49}\\ 
		& &\ours&\textbf{23.51}&\textbf{11.04}&\textbf{27.35}&\textbf{83.41}&60.53\\\bottomrule
    \end{tabular}
    \caption{\emph{Interpretability metrics} of \Ours \vs baseline \gap for different networks and interpretability methods on ImageNet.}
    \label{tab:intrecon-all}
    \vspace*{-3mm}
\end{table}
In \autoref{tab:intrecon-all}, we compare the interpretability properties when using our \ours \vs~\gap. In the appendix we provide comparisons with more models and datasets. We observe that \Ours provides consistent improvements over \gap in terms of AD, AG, AI and I metrics, while performing lower on D. Deletion (D) has raised concerns in previous works \cite{chefer2021transformer, zhang2023opti}. As (D) gradually blackens pixels, \emph{out-of-distribution} data~\cite{gomez2022metrics, hase2021outofdistribution, qiu2021resisting} is produced, possibly introducing bias~\cite{rong2022consistent}.
Moreover, non-spread attributions tend to perform better \cite{zhang2023opti}, which is likely the reason for lower performance.

\section{Conclusion}
In this work we observe that attention-based pooling in transformers is similar if not the same as forming a class agnostic CAM-based attribution. Based on this observation, we build upon this representation to mask features prior to the classification layers of a model, enhancing interpretability of existing image recognition models using \gap. Our method improves interpretability metrics while maintaining recognition performance.

\section{Acknowledgement}

This work has received funding from the UnLIR ANR
project (ANR-19-CE23-0009), the Excellence Initiative of Aix-Marseille Universite - A*Midex, a French
“Investissements d'Avenir programme” (AMX-21-IET-017). Part of this work was performed using
HPC resources from GENCI-IDRIS (Grant 2020-AD011011853 and 2020-AD011013110)

{
    \small
    \bibliographystyle{ieeenat_fullname}
    \bibliography{egbib}
}

\appendix
\newpage
\clearpage
\appendix
\section{More on the connection between Attention and CAM}
\begin{figure}[H]
    \centering
    \includegraphics[width=.475\textwidth]{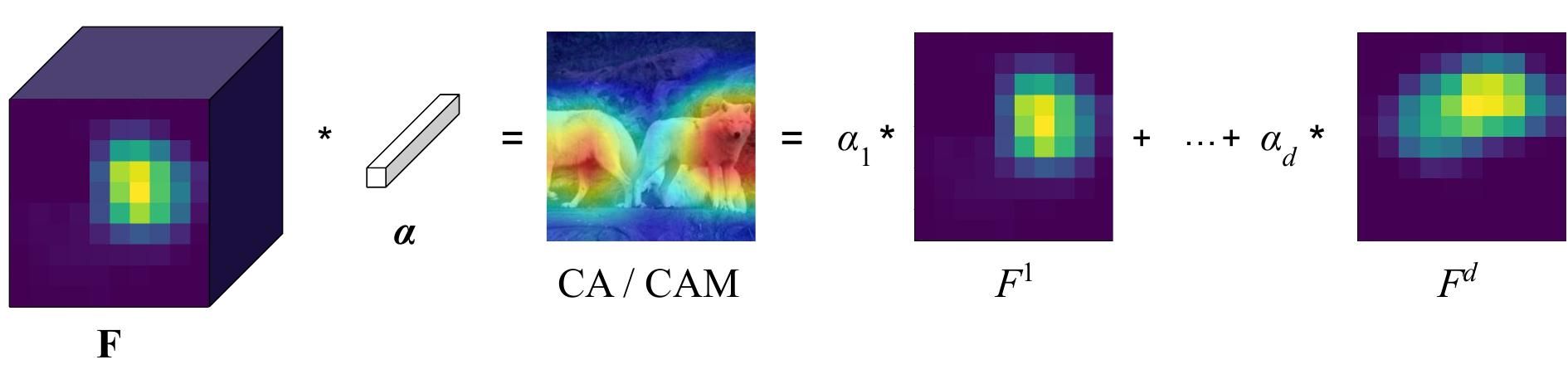}
    \caption{\textbf{Visualization of eq.~\eq{connection}.} On the left, a feature tensor $\vF \in \real^{w \times h \times d}$ is multiplied by the vector $\valpha \in \real^d$ in the channel dimension, like in $1 \times 1$ convolution, where $w \times h$ is the spatial resolution and $d$ is the number of channels. This is \emph{cross attention} (CA) \cite{dosovitskiy2020image} between the query $\valpha$ and the key $\vF$. On the right, a linear combination of feature maps $F^1, \dots, F^d \in \real^{w \times h}$ is taken with weights $\alpha_1, \dots, \alpha_d$. This is a \emph{class activation mapping} (CAM) \cite{zhou2016learning} with class agnostic weights. Eq.~\eq{connection} expresses the fact that these two quantities are the same, provided that $\valpha = (\alpha_1, \dots, \alpha_d)$ and $\vF$ is reshaped as $F = (\vf^1 \dots \vf^d) \in \real^{p \times d}$, where $p = wh$ and $\vf^k = \vect(F^k) \in \real^{p}$ is the vectorized feature map of channel $k$.}
    \label{fig:connection}
    
\end{figure}

Following the explanation of Cross-Attention acting as a class agnostic version of CAM demonstrated in \autoref{subsec:motiv}, we provide a visual explanation of this connection in \autoref{fig:connection}.
\section{More on experimental setup}
\paragraph{Implementation details}
Following the training recipes from the pytorch models\footnote{https://github.com/pytorch/vision/tree/main/references/classification}, we choose the ResNet protocol given its simplicity. Thus, we train over 90 epochs with SGD optimizer with momentum 0.9 and weight decay $10^{-4}$. We start our training with a learning rate of 0.1 and decrease it every 30 epochs by a factor of 10. Our models are trained on 8 V100 GPUs with a batch size 32 per GPU,  thus global batch size 256.
We follow the same protocol for both ResNet and ConvNeXt, though a different protocol might lead to improvements on ConvNeXt.
\section{More Visualizations}
\begin{figure*}[t]
\scriptsize
\centering
\setlength{\tabcolsep}{1.3pt}
     \begin{tabular}{cccccccc}
           \mc{2}{Corridor}&\mc{2}{Greenhouse}&\mc{2}{Pool Inside}&\mc{2}{Wine Cellar}\\
           Input image&Raw Attention&Input image&Raw Attention&Input image&Raw Attention&Input image&Raw Attention\\
           \includegraphics[width=0.12\textwidth,height=0.08\textwidth]{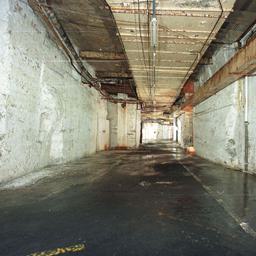}&
           \includegraphics[width=0.12\textwidth,height=0.08\textwidth]{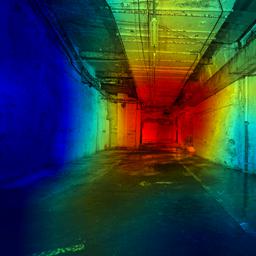}&
           \includegraphics[width=0.12\textwidth,height=0.08\textwidth]{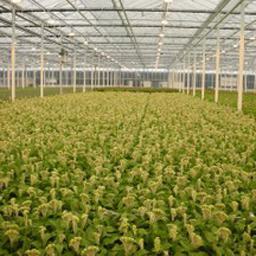}&
           \includegraphics[width=0.12\textwidth,height=0.08\textwidth]{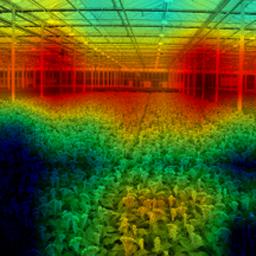}&
           \includegraphics[width=0.12\textwidth,height=0.08\textwidth]{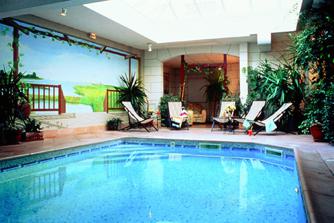}&
           \includegraphics[width=0.12\textwidth,height=0.08\textwidth]{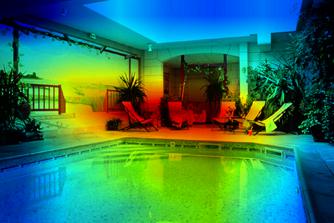}&
           \includegraphics[width=0.12\textwidth,height=0.08\textwidth]{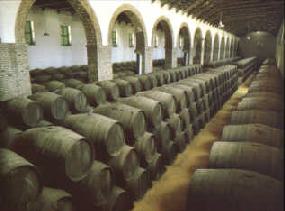}&
           \includegraphics[width=0.12\textwidth,height=0.08\textwidth]{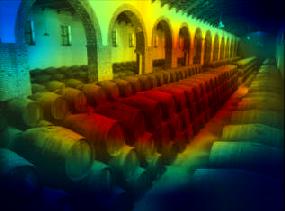}\\
           
           \includegraphics[width=0.12\textwidth,height=0.08\textwidth]{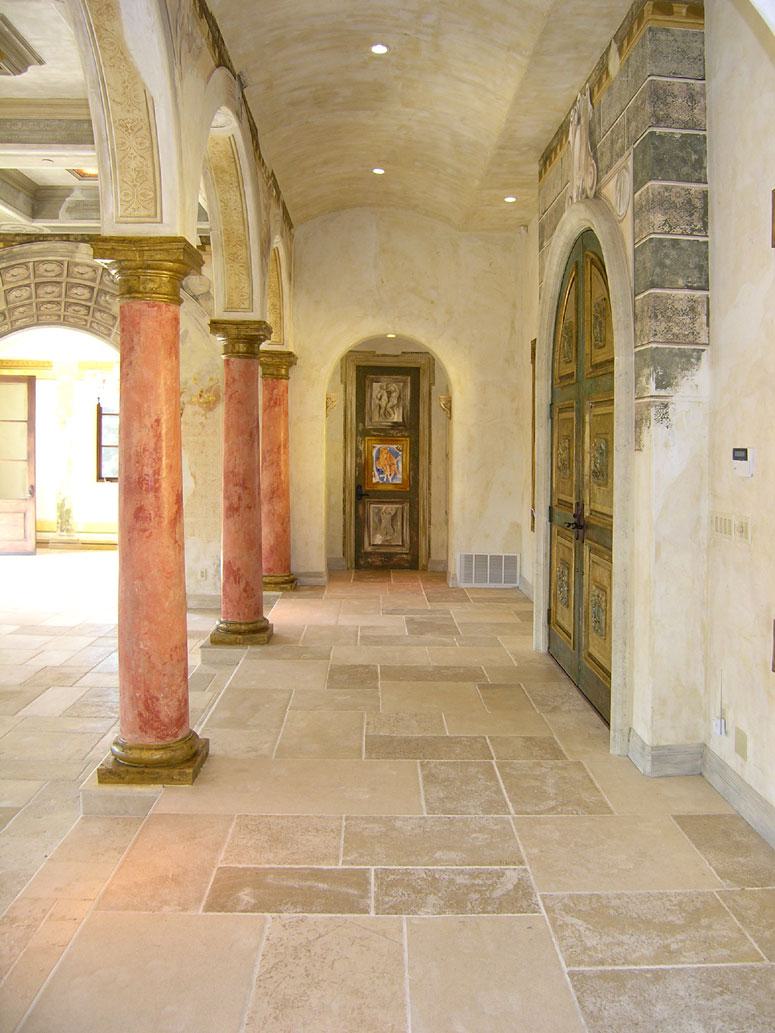}&
           \includegraphics[width=0.12\textwidth,height=0.08\textwidth]{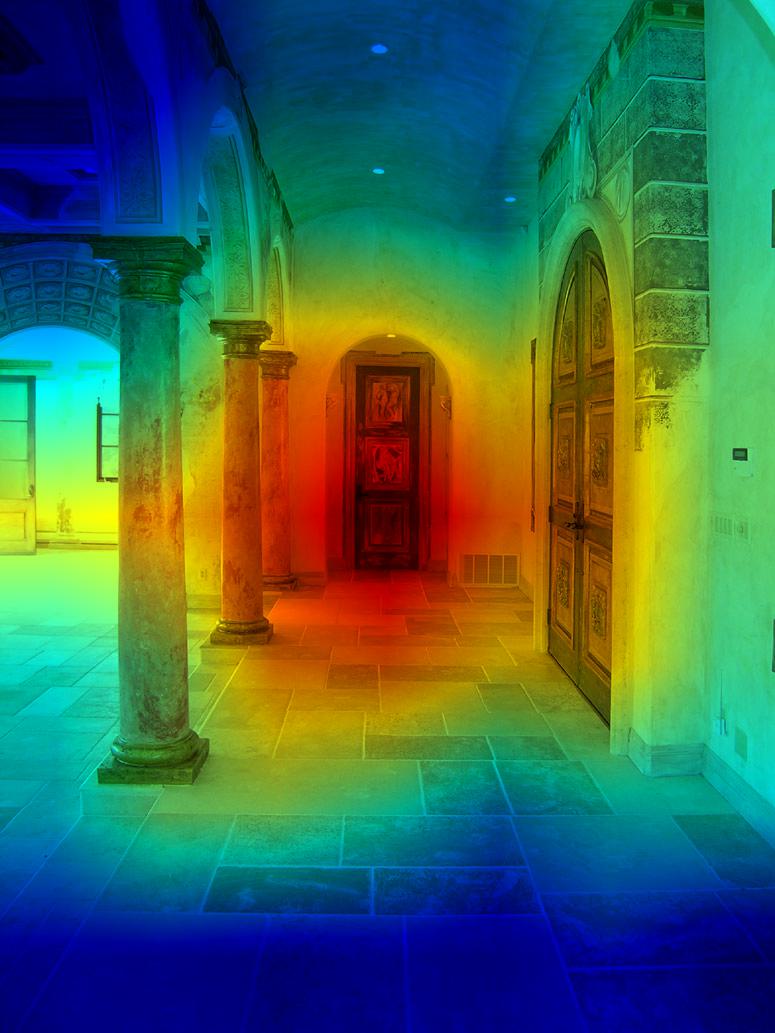}&
           \includegraphics[width=0.12\textwidth,height=0.08\textwidth]{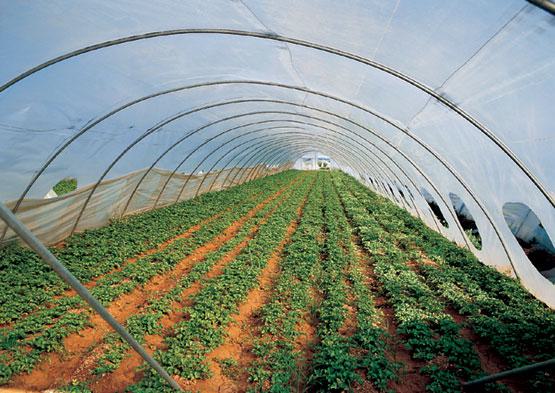}&
           \includegraphics[width=0.12\textwidth,height=0.08\textwidth]{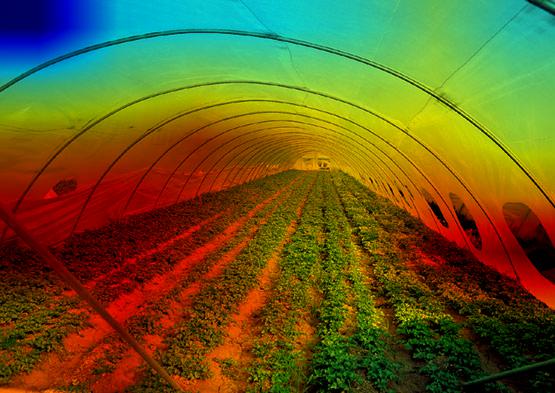}&
           \includegraphics[width=0.12\textwidth,height=0.08\textwidth]{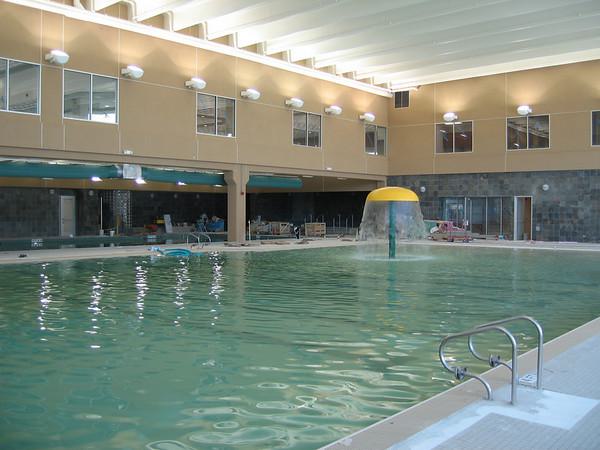}&
           \includegraphics[width=0.12\textwidth,height=0.08\textwidth]{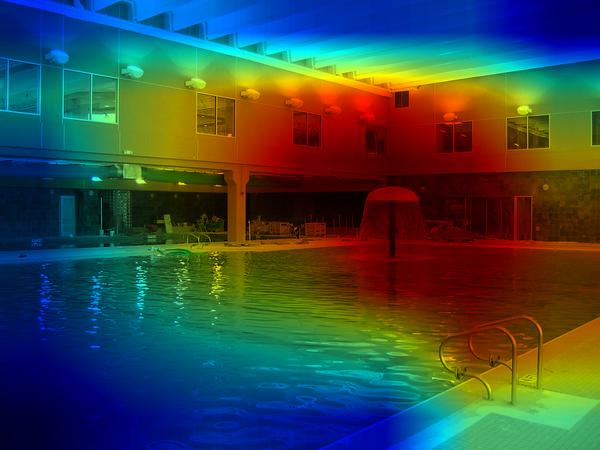}&
           \includegraphics[width=0.12\textwidth,height=0.08\textwidth]{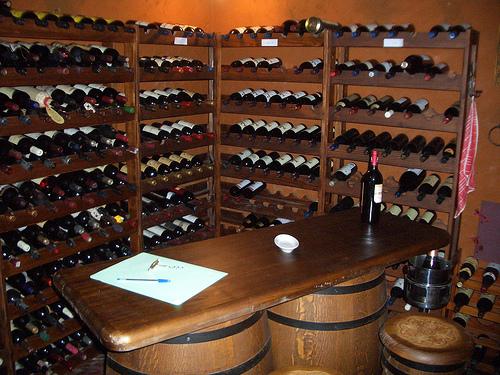}&
           \includegraphics[width=0.12\textwidth,height=0.08\textwidth]{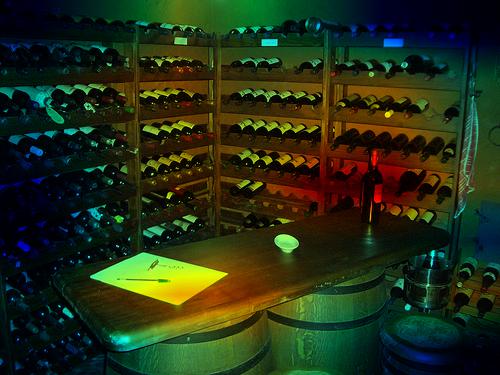}\\
           
           \includegraphics[width=0.12\textwidth,height=0.08\textwidth]{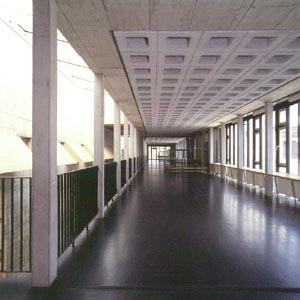}&
           \includegraphics[width=0.12\textwidth,height=0.08\textwidth]{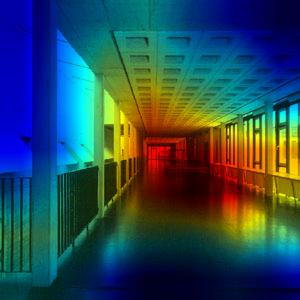}&
           \includegraphics[width=0.12\textwidth,height=0.08\textwidth]{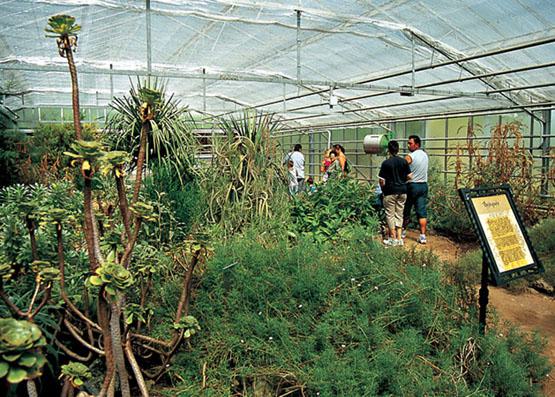}&
           \includegraphics[width=0.12\textwidth,height=0.08\textwidth]{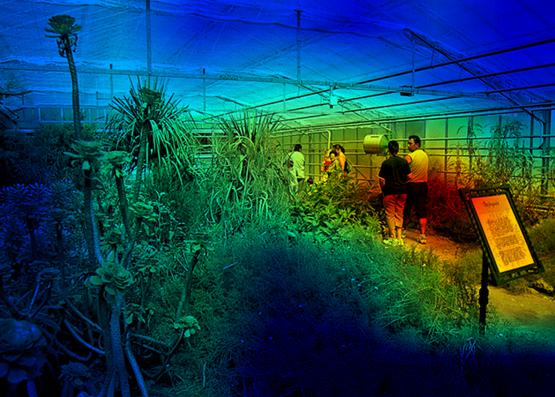}&
           \includegraphics[width=0.12\textwidth,height=0.08\textwidth]{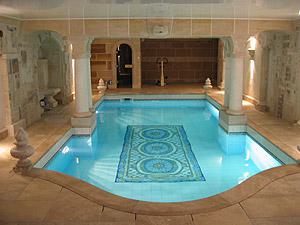}&
           \includegraphics[width=0.12\textwidth,height=0.08\textwidth]{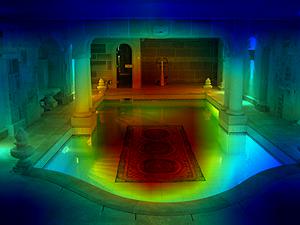}&
           \includegraphics[width=0.12\textwidth,height=0.08\textwidth]{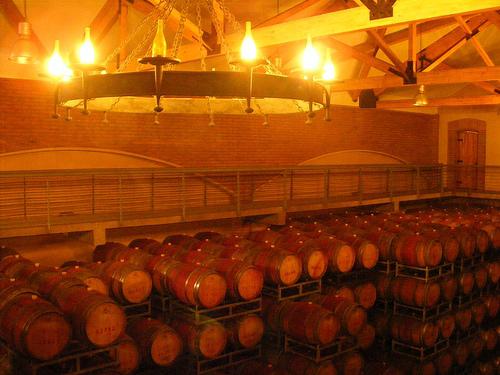}&
           \includegraphics[width=0.12\textwidth,height=0.08\textwidth]{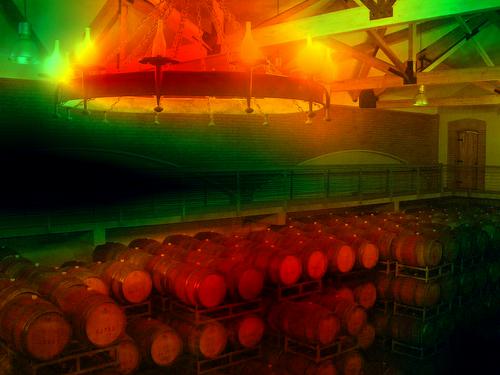}\\
           
           \includegraphics[width=0.12\textwidth,height=0.08\textwidth]{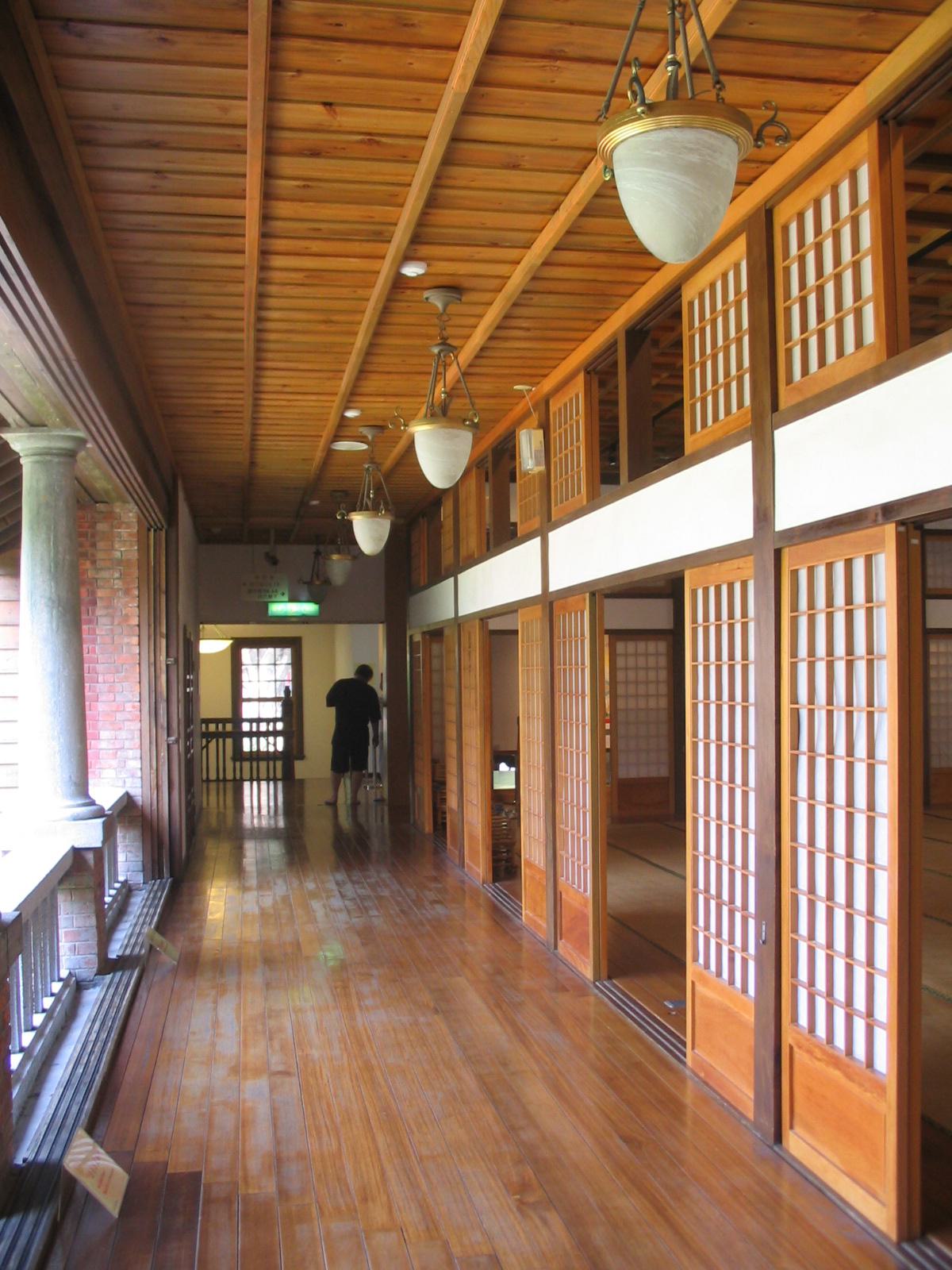}&
           \includegraphics[width=0.12\textwidth,height=0.08\textwidth]{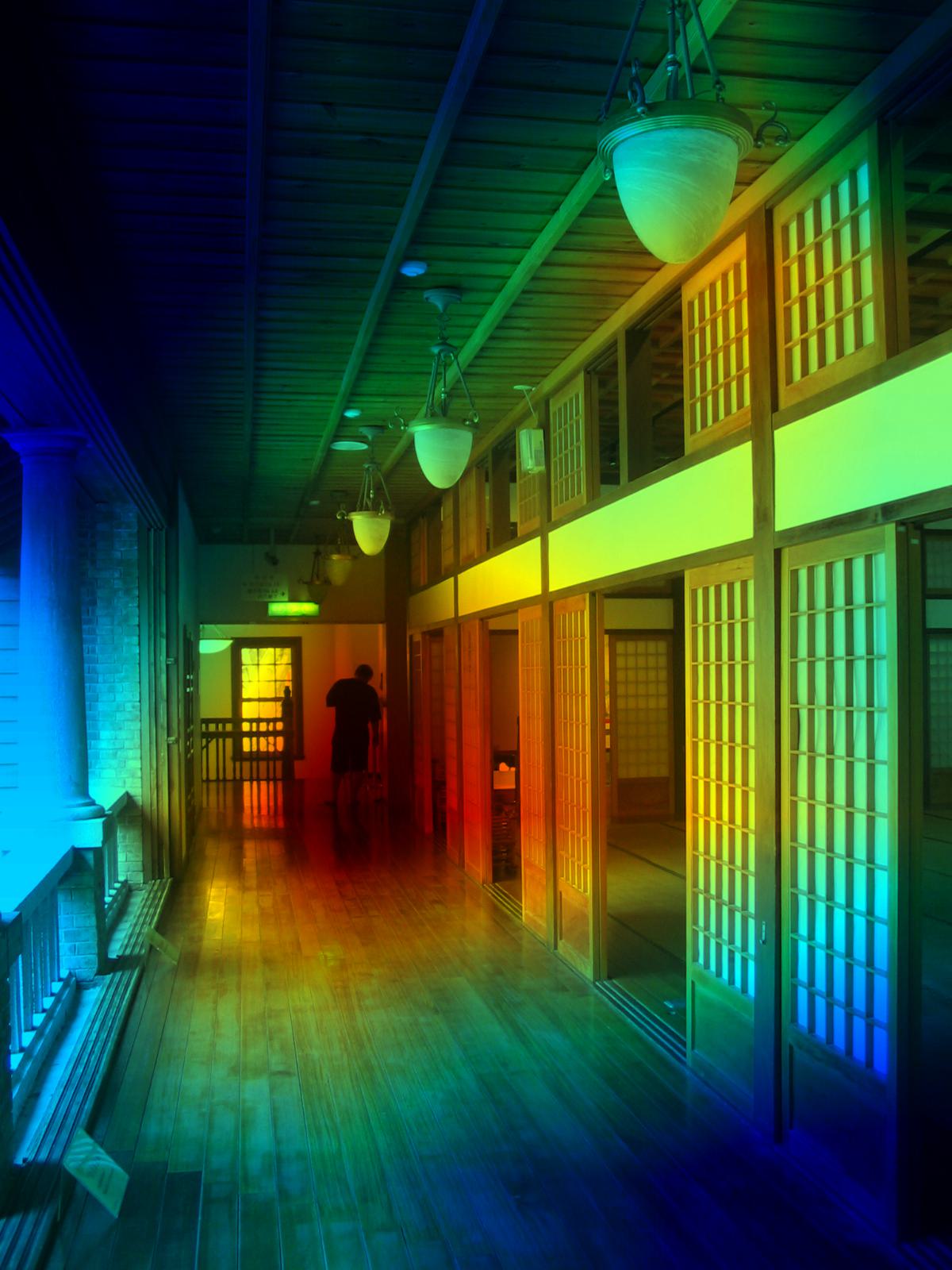}&
           \includegraphics[width=0.12\textwidth,height=0.08\textwidth]{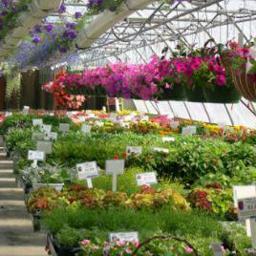}&
           \includegraphics[width=0.12\textwidth,height=0.08\textwidth]{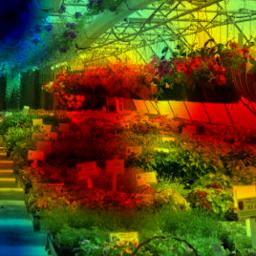}&
           \includegraphics[width=0.12\textwidth,height=0.08\textwidth]{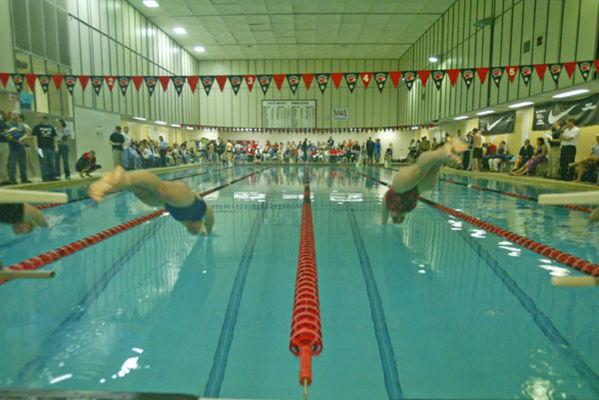}&
           \includegraphics[width=0.12\textwidth,height=0.08\textwidth]{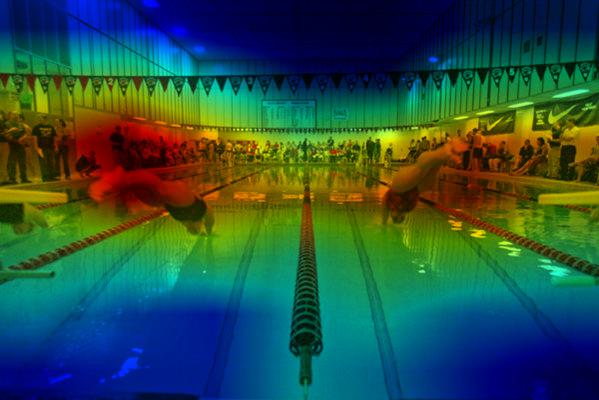}&
           \includegraphics[width=0.12\textwidth,height=0.08\textwidth]{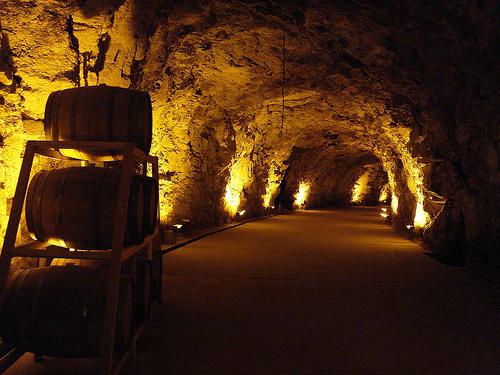}&
           \includegraphics[width=0.12\textwidth,height=0.08\textwidth]{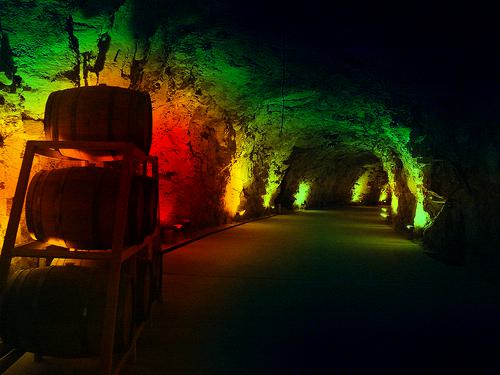}\\              
    \end{tabular}
    \vspace{3pt}
    \caption{Raw attention maps obtained from our \Ours on images of the MIT 67 Scenes dataset~\cite{quattoni2009recognizing} on classes that do not exist in ImageNet. The network sees them at inference for the first time.}
    \label{fig:enter-label}
\end{figure*}
In addition, \autoref{fig:enter-label} shows examples of images from the MIT 67 Scenes dataset~\cite{quattoni2009recognizing} along with raw attention maps obtained by \Ours. These images come from four classes that do not exist in ImageNet and the network sees them at inference for the first time. Nevertheless, the attention maps focus on objects of interest in general.
\section{More Architectures}
Table \autoref{tab:intrecon-morenets} presents interpretability metrics for both ResNet18 and ConvNeXt-S. Complementary experiments are reported on \autoref{tab:pascal} for CUB and Pascal VOC for ResNet 50.
\begin{table}[H]
    \centering
    \scriptsize
    \setlength{\tabcolsep}{2.5pt}
    \renewcommand{\arraystretch}{1.15}
    \begin{tabular}{llcccccc}
        \toprule
        \Th{Network}&\Th{Attribution}&\Th{Pooling}&\Th{AD$\downarrow$}&\Th{AG$\uparrow$}&\Th{AI$\uparrow$}&\Th{I$\uparrow$}&\Th{D$\downarrow$}\\\hline
	    \mr{7}{\Th{ResNet-18}}&\mr{2}{Grad-CAM}&\gap&17.64&12.73&41.21&63.13&\textbf{10.66}\\ %
		& &\ours&\textbf{16.99}&\textbf{17.22}&\textbf{44.95}&\textbf{65.94}&10.68\\\cmidrule{2-8} %
		& \mr{2}{Grad-CAM++}&\gap&19.05&11.16&37.99&62.80&\textbf{10.75}\\ %
		& &\ours&\textbf{19.02}&\textbf{14.76}&\textbf{40.82}&\textbf{65.53}&10.82\\\cmidrule{2-8} %
		& \mr{2}{Score-CAM}&\gap&13.64&12.98&44.53&62.56&\textbf{11.37}\\ %
		& &\ours&\textbf{11.53}&\textbf{18.12}&\textbf{50.32}&\textbf{65.33}&11.51\\\midrule %
	    \mr{7}{\Th{ConvNeXt-S}}&\mr{2}{Grad-CAM}&\gap&42.99&1.69&12.60&48.42&\textbf{30.12}\\ 
		& &\ours&\textbf{22.09}&\textbf{14.91}&\textbf{32.65}&\textbf{84.82}&43.02\\\cmidrule{2-8} 
		& \mr{2}{Grad-CAM++}&\gap&56.42&1.32&10.35&48.28&\textbf{33.41}\\ 
		& &\ours&\textbf{51.87}&\textbf{9.40}&\textbf{20.55}&\textbf{84.28}&52.58\\\cmidrule{2-8} 
		& \mr{2}{Score-CAM}&\gap&74.79&1.29&10.10&47.40&\textbf{38.21}\\ 
		& &\ours&\textbf{64.21}&\textbf{8.81}&\textbf{18.96}&\textbf{82.92}&57.46\\\bottomrule
    \end{tabular}
    \caption{of \Ours \vs baseline \gap for more networks and interpretability methods on ImageNet.}
    \label{tab:intrecon-morenets}
\end{table}
\begin{table}[H]
    \centering
    \scriptsize
    \setlength{\tabcolsep}{4pt}
    \begin{tabular}{llccccc}\toprule
        \mc{7}{\textbf{\Th{CUB-200-2011 - ResNet-50}}}\\\midrule
        &\Th{Pooling}&\mc{2}{}&\mc{2}{}&\Th{Acc$\uparrow$}\\\midrule
        &\gap&\mc{2}{}&\mc{2}{}&76.96\\
        &\ours&\mc{2}{}&\mc{2}{}&75.90\\\midrule

        \mc{7}{\Th{Interpretability Metrics}}\\\midrule
        \Th{Method}&\Th{Pooling}&AD$\downarrow$&AG$\uparrow$&AI$\uparrow$&I$\uparrow$&D$\downarrow$\\\midrule
        \mr{2}{Grad-CAM}&\gap&10.87&10.29&45.81&65.71&\textbf{6.17}\\
            &\ours&\textbf{10.44}&\textbf{17.61}&\textbf{53.54}&\textbf{74.60}&6.56\\\midrule
        \mr{2}{Grad-CAM++}&\gap&11.35&9.68&44.32&65.64&\textbf{5.92}\\
        &\ours&\textbf{11.01}&\textbf{16.50}&\textbf{51.63}&\textbf{74.64}&6.21\\\midrule
        \mr{2}{Score-CAM}&\gap&9.05&10.62&48.90&65.58&5.94\\
        &\ours&\textbf{6.37}&\textbf{19.50}&\textbf{60.41}&\textbf{74.22}&\textbf{2.14}\\
        \midrule
        \midrule
        \mc{7}{\textbf{\Th{Pascal VOC 2012 - ResNet-50}}}\\\midrule
        &\Th{Pooling}&\mc{2}{}&\mc{2}{}&\Th{mAP$\uparrow$}\\\midrule
        &\gap&\mc{2}{}&\mc{2}{}&78.32\\
        &\ours&\mc{2}{}&\mc{2}{}&78.35\\\midrule
        \mc{7}{\Th{Interpretability Metrics}}\\\midrule
        \Th{Method}&\Th{Pooling}&AD$\downarrow$&AG$\uparrow$&AI$\uparrow$&I$\uparrow$&D$\downarrow$\\\midrule%
        \mr{2}{Grad-CAM}&\gap&\textbf{12.61}&9.68&27.88&\textbf{89.10}&59.39\\
        &\ours&12.77&\textbf{15.46}&\textbf{34.53}&88.53&\textbf{59.16}\\\midrule
        \mr{2}{Grad-CAM++}&\gap&\textbf{12.25}&9.68&27.62&\textbf{89.34}&54.23\\
        &\ours&12.28&\textbf{16.76}&\textbf{34.87}&89.02&\textbf{53.34}\\\midrule
        \mr{2}{Score-CAM}&\gap&14.8&6.76&36.41&71.10&\textbf{39.95}\\
        &\ours&\textbf{10.96}&\textbf{21.35}&\textbf{43.82}&\textbf{89.21}&51.44\\\bottomrule
    \end{tabular}
    \caption{Accuracy, respectively mean Average Precision, and interpretability metrics of \Ours \vs baseline \gap for ResNet-50 on CUB and Pascal dataset.}
    \label{tab:pascal}
\end{table}

Results on CUB in \autoref{tab:pascal} show that our \Ours consistently provides improvements when the model is fine-tuned on a smaller fine-grained dataset.
\section{Ablation Experiments}
We conduct ablation experiments on ResNet50 because of its modularity and ease of modification. We investigate the effect of the cross attention block design, the placement of the \Ours relative to the backbone network.
\paragraph{Cross attention block design}
Following transformers~\cite{NIPS2017_3f5ee243,dosovitskiy2020image}, it is possible to add more layers in the cross attention block. We consider a variant referred to as \PO, which uses linear projections $W_\ell^K, W_\ell^V \in \real^{d_\ell \times d_\ell}$ on the key and value
\begin{equation}
	\ca_\ell(\vq_\ell, F_\ell) \defn (F_\ell W^V_\ell)\tran h_\ell(F_\ell W^K_\ell \vq_\ell)\in \real^{d_\ell},
\label{eq:proj_ca}
\end{equation}
while ~\eq{qk-layer} remains.
\begin{table}[H]
\centering
\scriptsize
\begin{tabular}{lcc}\toprule
	\Th{Block Type}&\Th{$\#$Params}&\Th{Accuracy}\\\midrule
	\our&6.96M&74.70\\
	\PO&18.13M&74.41\\\bottomrule
\end{tabular}
\caption{\emph{Different cross attention block design for \Ours.} Classification accuracy and parameters using ResNet-50 on ImageNet. \Th{$\#$Param}: parameters of \Ours only.}
\label{tab:dif_streams}
\end{table}
Results are reported in \autoref{tab:dif_streams}. We observe that the stream made of vanilla CA blocks ~\eq{CA} offers slightly better accuracy than projections, while having less parameters. We also note that most of the computation takes place in the last residual stages, where the channel dimension is the largest. To keep our design simple, we choose the vanilla solution without projections ~\eq{CA} by default.
\paragraph{\Ours placement}
\label{ab:placement}
To validate the design of \Ours, we measure the effect of its depth on its performance \vs the baseline \gap in terms of both classification accuracy / number of parameters and classification metrics for interpretability. In particular, we place the stream in parallel to the network $f$, starting at stage $\ell$ and running through stage $L$, the last stage of $f$, where $0 \le \ell \le L$. Results are reported in \autoref{tab:intrecog-resnet}.
\begin{table}[H]
\centering
\scriptsize
\setlength{\tabcolsep}{4pt}
\begin{tabular}{lcccccc}\toprule
	\mc{7}{\Th{Accuracy and Parameters}}\\\midrule
	&\Th{Placement}&\mc{2}{\Th{CLS dim}}&\mc{2}{\Th{\#Param}}&\Th{Acc$\uparrow$}\\\midrule
	
	&$S_0-S_4$&\mc{2}{$64$}  &\mc{2}{6.96M}&\textbf{74.70}\\  
	&$S_1-S_4$&\mc{2}{$256$} &\mc{2}{6.95M}&74.67\\           
	&$S_2-S_4$&\mc{2}{$512$} &\mc{2}{6.82M}&74.67\\           
	&$S_3-S_4$&\mc{2}{$1024$}&\mc{2}{6.29M}&74.67\\           
	&$S_4-S_4$&\mc{2}{$2048$}&\mc{2}{4.20M}&74.63\\\midrule   
	
	\mc{7}{\Th{Interpretability Metrics}}\\\midrule
	\Th{Method}&\Th{Placement}&\Th{AD$\downarrow$}&\Th{AG$\uparrow$}&\Th{AI$\uparrow$}&\Th{I$\uparrow$}&\Th{D$\downarrow$}\\\midrule
	
	\mr{5}{\Th{Grad-CAM}}&$S_0-S_4$&\textbf{12.54}&\textbf{22.67}&48.56&75.53&13.50\\ 
		&$S_1-S_4$&12.69&22.65&48.31&75.53&13.41\\ 
		&$S_2-S_4$&\textbf{12.54}&21.67&\textbf{48.58}&75.54&13.50\\ 
		&$S_3-S_4$&12.69&22.28&47.89&\textbf{75.55}&13.40\\ 
		&$S_4-S_4$&12.77&20.65&47.14&74.32&\textbf{13.37}\\\midrule 
		
	\mr{5}{\Th{Grad-CAM++}}&$S_0-S_4$&13.99&19.29&44.60&75.21&13.78\\ 
		&$S_1-S_4$&13.99&19.29&44.62&75.21&13.78\\ 
		&$S_2-S_4$&13.71&\textbf{19.90}&\textbf{45.43}&75.34&13.50\\ 
		&$S_3-S_4$&13.69&19.61&45.04&\textbf{75.36}&13.50\\ 
		&$S_4-S_4$&\textbf{13.67}&18.36&44.40&74.19&\textbf{13.30}\\\midrule 
		
	\mr{5}{\Th{Score-CAM}}&$S_0-S_4$&\textbf{7.09}&23.65&54.20&74.91&14.68\\ 
		&$S_1-S_4$&\textbf{7.09}&23.65&54.20&74.92&14.68\\ 
		&$S_2-S_4$&\textbf{7.09}&\textbf{23.66}&\textbf{54.21}&74.91&14.68\\ 
		&$S_3-S_4$&7.74&23.03&52.92&\textbf{74.97}&14.65\\ 
		&$S_4-S_4$&7.52&19.45&50.45&74.19&\textbf{14.46}\\\bottomrule 
\end{tabular}
\caption{\emph{Effect of stream placement} on accuracy, parameters and interpretability metrics  for ResNet-50 on ImageNet. $S_\ell-S_L$: \Ours runs from stage $\ell$ to $L$ (last); \Th{$\#$Param}: parameters of \Ours only.}
\label{tab:intrecog-resnet}
\end{table}
From the interpretability metrics as well as accuracy, we observe that stream configurations that allow for iterative interaction with the network features obtain the best performance, although the effect of stream placement is small in general. In many cases, the lightest stream of only one cross attention block ($S_4-S_4$) is inferior to options allowing for more interaction. 
Since starting the stream at early stages has little effect on the number of parameters and performance is stable, we choose to start the stream in the first stage ($S_0-S_4$) by default.

\paragraph{Class-specific CLS}
As discussed in \autoref{subsec:CA-base}, the formulation of single-query cross attention as a CAM-based saliency map ~\eq{sal} is class agnostic (single channel weights $\alpha_k$), whereas the original CAM formulation ~\eq{sal} is class specific (channel weights $\alpha_k^c$ for given class of interest $c$). 
Here we consider a class specific extension of \Ours using one query vector per class. 
In particular, the stream is initialized by one learnable parameter $\vq_0^c$ per class $c$, but only one query (\cls token) embedding is forwarded along the stream. At training, $c$ is chosen according to the target class label, while at inference, the class predicted by the baseline classifier is used instead.
\begin{table}[H]
\centering
\scriptsize
\setlength{\tabcolsep}{4pt}
    \begin{tabular}{llccccc}\toprule                    
	   \mc{7}{\Th{Accuracy and Parameters}}\\\midrule
	   &\Th{Representation}&\mc{2}{}&\mc{2}{\Th{\#Param}}&\Th{Acc$\uparrow$}\\\midrule
		&Class agnostic&\mc{2}{}&\mc{2}{32.53M}&74.70\\
		&Class specific&\mc{2}{}&\mc{2}{32.59M}&74.68\\\midrule	
	   \mc{7}{\Th{Interpretability Metrics}}\\\midrule
    \Th{Method}&Th{Representation}&AD$\downarrow$&AG$\uparrow$&AI$\uparrow$&I$\uparrow$&D$\downarrow$\\\midrule    
       \mr{2}{Grad-CAM}&Class agnostic&12.54&22.67&48.56&75.53&13.50\\
	   &Class specific&12.53&22.66&48.58&75.54&13.50\\\midrule
	   \mr{2}{Grad-CAM++}&Class agnostic&13.99&19.29&44.60&75.21&13.78\\
	     &Class specific&13.99&19.28&44.62&75.20&13.78\\\midrule
	   \mr{2}{Score-CAM}&Class agnostic&7.09&23.65&54.20&74.91&14.68\\
	   &Class specific&7.08&23.64&54.15&74.99&14.53\\\bottomrule
    \end{tabular}
    \vspace{3pt}
    \caption{\emph{Effect of class agnostic \vs class specific representation} on accuracy, parameters and interpretability metrics of \Ours for ResNet-50 and different interpretability methods on ImageNet. \Th{$\#$Param}: parameters of \Ours only.}
    \label{tab:TokenvMatrix}
\end{table}

Results are reported in \autoref{tab:TokenvMatrix}. We observe that the class specific representation for \Ours provides no improvement over the class agnostic representation, despite the additional complexity and parameters. We thus choose the class agnostic representation by default. The class specific approach is similar to [50] in being able to generate class specific attention maps, although no fine-tuning is required in our case.



\end{document}